\documentclass[twoside,11pt]{article} 
\usepackage{jmlr2e} 
\usepackage{color}
\usepackage{graphics}
\usepackage{makecell}
\usepackage{graphicx}
\usepackage{float}
\usepackage{amsmath}
\usepackage{subfig}
\usepackage{longtable}
\usepackage{bm}
\usepackage [autostyle, english = american]{csquotes}
\usepackage{url}
\usepackage{hyperref}
\MakeOuterQuote{"}

\ShortHeadings{Can Intelligent Hyperparameter Selection Improve Resistance to Adversarial Examples?}{Burkard and Lagesse}
\firstpageno{1}

\begin{document}
 
 \title{Can Intelligent Hyperparameter Selection Improve Resistance to
 	Adversarial Examples?}
 \author{\name Cody Burkard \email cburkard@uw.edu \\
 	\addr Computing and Software Systems\\
 	University of Washington Bothell\\
 	Bothell, WA
 	\AND
 	\name Brent Lagesse \email lagesse@uw.edu \\
 	\addr Computing and Software Systems\\
University of Washington Bothell\\
Bothell, WA}

\maketitle

\begin{abstract}%
	Convolutional Neural Networks and Deep Learning classification systems in general have been shown to be vulnerable to attack by specially crafted data samples that appear to belong to one class but are instead classified as another, commonly known as \textit{adversarial examples}. A variety of attack strategies have been proposed to craft these samples; however, there is no standard model that is used to compare the success of each type of attack. Furthermore, there is no literature currently available that evaluates how common hyperparameters and optimization strategies may impact a model's ability to  resist these samples. This research bridges that lack of awareness and provides a means for the selection of training and model parameters in future research on evasion attacks against convolutional neural networks. The findings of this work indicate that the selection of model hyperparameters does impact the ability of a model to resist attack, although they alone cannot prevent the existence of adversarial examples.
 
\end{abstract}


\section{Introduction}
\label{sec:introduction}

The increasing ability of machine learning systems to automatically perform complicated tasks such as image classification has changed the way that our society makes decisions about the world around us. An MIT technology review \cite{MIT_review} found that 60\% of companies surveyed had already implemented some type of machine learning strategy, while 18\% indicated that they plan to implement a strategy within the next two years. Some high profile examples of companies that use these systems include Google's use of GoogleNet to classify images, Facebook's DeepFace for performing facial recognition, Google's AlphaGo \cite{silver_mastering_2016} system for playing (and winning) the game of Go, and even Nvidia's Dave-2 system for training self-driving cars \cite{bojarski_end_2016}. All of these systems, as well as many others, are largely dependent on convolutional neural networks(CNNs). This algorithm is a specialized type of machine learning algorithm that can extract spatial relationships between its input features. In recent years, CNNs have gained a significant amount of traction with software companies, largely due to its enormous success in image processing tasks.

Machine learning algorithms, including CNNs, have been shown to be vulnerable to manipulation in an adversarial environment. This vulnerability presents a new attack vector on the many systems that deploy these algorithms. A system employing machine learning to automate decisions may be vulnerable to certain types of attack that either cause it to misclassify legitimate data, or trick it to misclassify illegitimate data. The later has been explored extensively on multiple types of learning systems, especially image recognition systems that utilize CNNs\cite{szegedy_intriguing_2013,goodfellow_explaining_2014,nguyen_deep_2015,papernot_limitations_2015,carlini_towards_2016,moosavi-dezfooli_deepfool:_2016}. This is called an evasion attack, and in the context of CNNs, the resulting malicious sample is called an \textit{adversarial example}. Currently, although many attacks have been constructed, little is known about whether common algorithmic decisions for these algorithms, such as hyperparameter choice and optimization strategy, provide any type of security against these attacks.

While the prospect of an evasion attack against commonly studied classification systems such as spam detection may not seem ominous, consider this attack in the context of self-driving car. Regardless of the physical security of a car or the security of its network, it is not a safe system if an attacker can cause the car to run a stop sign simply by drawing on the sign with a marker and causing it to be classified as a no parking sign. As CNNs are deployed in more cyber-physical systems, the risk of an evasion attack is no longer the loss of emails or information; it is the loss of human lives.

Current literature briefly explores the ability of a CNN's hyperparameters such as regularization to defend against adversarial examples, but only states that it is not a valid defense without any explanation\cite{goodfellow_explaining_2014}. Beyond this assertion, there is a lack of understanding about how hyperparameters of a network impact its resistance to adversarial examples. The open question is; \textit{how do hyperparameters of a CNN impact its ability to fit the true function for the data in its environment?} This work hypothesizes that intelligent choice of network's hyperparameters is a necessary but not sufficient condition to help reduce the possible space for adversarial examples. The contributions of this work include a new type of analysis of the inherent vulnerabilities in a commonly used machine learning dataset, and a comparison of the impact of common hyperparameters on the security of a CNN. These contributions provide insight into the secure training of a CNN, and a basis for comparing the security of future models.

\subsection{Motivation}
\label{sec:motivation}

CNNs are becoming ubiquitous in many cutting edge technologies, such as health care, social media, and transportation. This is largely due to the image processing ability of CNNs for tasks such as recognizing skin diseases, or vision in self-driving cars. The capability of visually processing the world with CNNs is yielding new types of systems that are more integrated into our physical lives than ever before.  This integration into the physical world, as well as research into the integration of these algorithms into critical systems such as the electric grid\cite{2016arXiv160408382C}, is motivation enough to understand what attacks are possible on these systems and how they might be prevented or mitigated.

Previous works have focused on crafting adversarial examples in new ways as well as defending models against these attacks; however, they craft these samples on a single model with no point of comparison to models trained with different hyperparameters. Furthermore, there no insight is given as to why the attack was employed on the specific model, whether is was simplicity, accuracy, memory efficiency, etc. Without a thorough understanding of the difference in vulnerability between models trained with subtle differences, results for both attacks and defenses against these different models become difficult to compare. It is our hope that this work provides future research into adversarial examples with a basis for choosing the hyperparameters to test adversarial example crafting mechanisms and defenses against. Furthermore, a greater understanding about the impact of hyperparameters on the vulnerability of a model may also lead to the design of models that are more robust to adversarial influence.

This task is challenging because machine learning algorithms are designed to handle tasks where there does not exist a traditional algorithm that can provide a 100\% correct solution. However, without this knowledge it is difficult to determine how "wrong" a learned model's function is; in other words, given a set of all the possible samples of data of a certain type, how many of those samples are classified correctly by the learned model? This cannot be directly measured in a tractable amount of time, so to compare the security of different models a more creative approach must be taken.

As is true in performance based machine learning research, any results that are found have to be made on \textit{observable data}, or data that has previously been seen in the environment. In the case of this work, the observable data that can be used for comparison is previously constructed adversarial examples. Methods are available to generate new observable data through the use of adversarial example crafting techniques introduced by other work; however, these techniques do not fully represent the realm of all possible adversarial examples. Without a a true representation of the entire space of adversarial examples, the task of understanding how to measure and compare the robustness of multiple models is non-trivial. One potential method for this measurement is explored in this work, and the approach used for that measurement is described in section \ref{sec:methods}.

The remainder of this paper is organized as follows:  First, an introduction to Convolutional Neural Networks and the field of Adversarial Machine Learning. Following this, current attacks against CNNs are discussed in depth, along with current defense mechanisms. The reason that these attacks are possible is then discussed in detail, along with specific issues pertaining to data sets. The methodology of this work is then presented, followed by the results of the experimentation done in this work. Finally, this work ends with a discussion on the experimental results, the current ability to defend against adversarial examples and an overview of the contributions of this work.

\section{Background and Notation}
\label{sec:background}

In this section, we present an overview of convolutional neural networks. Some of the challenges of these algorithms are discussed, and finally an overview of attacks against classification systems such as CNNs is presented.

\subsection{Convolutional Neural Networks}

The success of Neural Networks when applied to computer vision is limited on large images due to the \textit{curse of dimensionality}; as the number of pixels in an image increase, the volume of the feature space for a dataset increases exponentially. Because of this, the amount of data needed to train the network to represent its feature space also increases exponentially. If the number of training samples used to represent this dataset is too small, the classifier only learns to separate samples based on features that are specific to the training data instead of features that truly represent the underlying classes that those samples belong to. 

In order to combat this constraint on learning with large images, Lecun\cite{lecun1989} proposed the use of a mathematical operation called a \textit{convolution} as a layer in neural networks, creating a special type of neural network called a Convolutional Neural Network, or CNN. In machine learning, this operation is useful for extracting features from grid-like data with important spatial relationships between input features in the grid. It has been shown to be useful for a variety of machine learning applications such as image processing, where images are commonly represented as a grid of pixels. In this section, the basic workflow and common operations in a CNN are described.

\subsubsection{Convolution Layer}

\begin{figure}[H]
\centering
\includegraphics[width=\textwidth,height=\textheight,keepaspectratio]{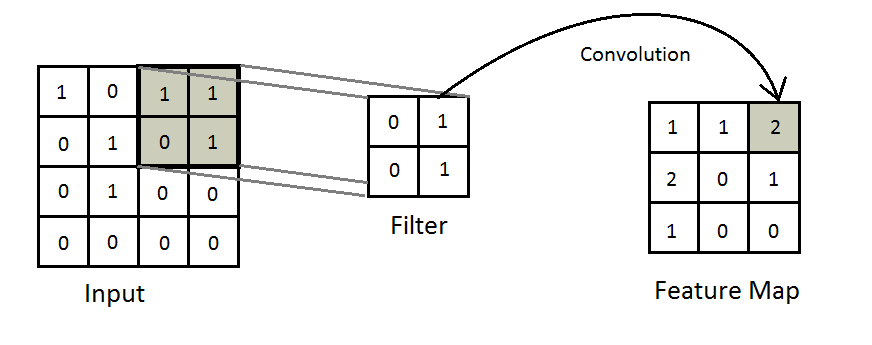}
\caption{Convolution Layer Example. The stride is 1, and the filter size is 2x2.}
\label{fig:convolution}
\end{figure}

The convolution layer in a CNN is primarily responsible for the extraction of spatial relationships between input features. This is done by iterating a smaller matrix of weights, called a \textit{filter}, over the grid-like input to the network to detect features of the input. The distance that the filter "slides" over the input during each iteration is referred to as \textit{stride}. At each iteration, the dot product of the filter and a \textit{slice} of the input that it overlaps is computed and used as a feature in an output matrix, and then the filter slides over the input according to the stride. Once a filter has iterated over the entire input, the output matrix that is constructed is referred to as a \textit{feature map}. For a visual example of this process, see figure \ref{fig:convolution}.

Note that during this process, the same filter weights are used on each slice of an input to extract features of a feature map. This is in stark contrast to a fully connected layer, in which each input feature has a weight for each neuron in the next layer. By using the same filter on each slice of input, the number of parameters in the network is dramatically reduced. This idea is referred to as \textit{parameter sharing}.

The size of the output of the convolution layer is dependent on the size of the input, the size of the filters, the stride used, and the number of filters used. The input, filter, and output of a convolution layer has three dimensions: width, height and depth. In the case of an image, the width and the height of the input would refer to the number of pixels wide and number of pixels high of that image. The depth of an image would refer to the number of color channels in the image, such as the red, green, and blue channels. In a grayscale image, there is only one color channel, so the depth is only one.

The width and height of a filter are parameters that can be modified, but the depth is always the same depth of the input depth. Smaller filters take more iterations to cover the entire input, so the number of features extracted is larger. On the other hand, larger strides can reduce the number of iterations it takes to cover the input, reducing the width and height of the output feature map. The depth of the feature map always represents the number of filters used to iterate over the input.

During backpropagation, the chain rule is used to compute the gradient of each feature in the convolution layer output. The filters that were used during the forward propagation of a samples are flipped by 180 degrees and the convolution operation is applied to the computed gradients to reconstruct a new input feature map. 

\subsubsection{Max-Pool Layer}

It is common practice to use a \textit{pooling} layer after some convolution layers to reduce feature dimensionality and the number of parameters in the network. This is sometimes referred to as \textit{down-sampling}. Pooling layers help to prevent overfitting and promote translation invariance, or the ability of a network to classify objects in different orientations in an input image. In this  section, max-pooling is discussed. There are many other types of pooling, however they are not used in this work so they are left out of this discussion.

\begin{figure}[H]
\centering
\includegraphics[width=\textwidth,height=\textheight,keepaspectratio]{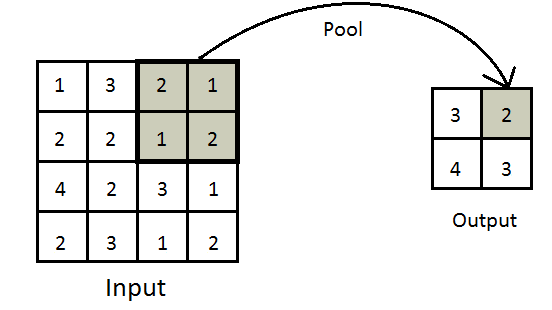}
\caption{Max-Pooling Example. The stride is 2, and the pool size is 2x2.}
\label{fig:max-pooling}
\end{figure}

Like the convolution layer, a max-pooling layer accepts a three dimensional input of some width, height, and depth. It includes two parameters, stride and pool filter size. Stride is used in the same way as it is in the convolution layer, however the filter does not have any weights and is instead an operation that finds the maximum value at each slice of an input. The output dimension has the same depth as the input, but has a smaller width and height depending on the size of the pool and stride size. This process is visualized in figure \ref{fig:max-pooling}

\subsubsection{Activation Layer}

In order to model data that is not linearly separable, non-linearity must be introduced to a network. In MLPs and CNNs, this comes in the form of a non-linear activation function that acts on the output of a neuron. A common activation function that is used is the Rectified Linear Unit(ReLU) activation function, which performs a threshold operation on each output value where any value less than zero is set to zero. Other common activation functions include sigmoid, tanh, and ELU.

\begin{figure}[H]
\centering
\includegraphics[width=\textwidth,height=\textheight,keepaspectratio]{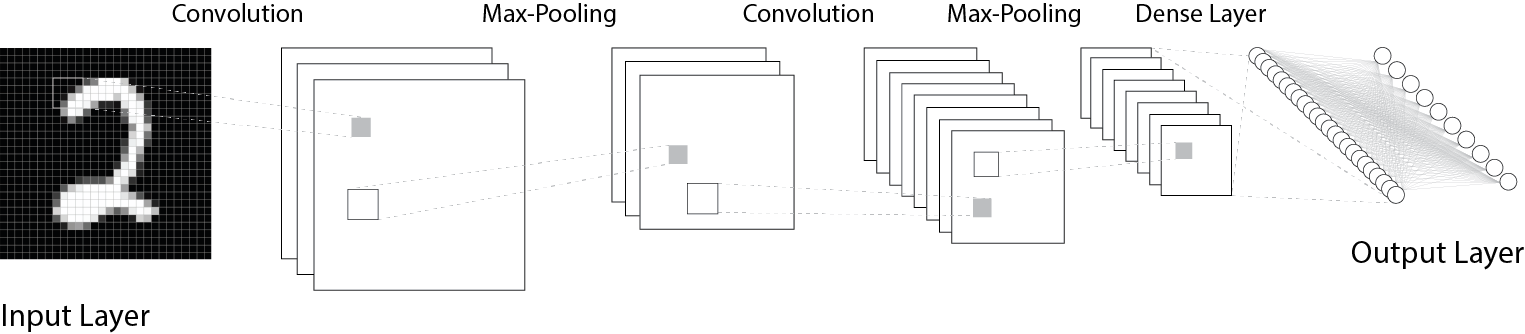}
\caption{CNN Architecture Example}
\label{fig:CNN}
\end{figure}

These layers are the basis of many modern image recognition systems. Many of these systems are inspired by the LeNet architecture proposed by \cite{lecun1998gradient}. This architecture found immense success on handwritten digit recognition by combining two convolution layers with down-sampling layers immediately following each, and is the basis of many modern systems such as DeepFace\cite{deepface} and GoogleNet\cite{googleNet}. Figure \ref{fig:CNN} provides a visual example of a LeNet Convolutional Neural Network architecture.

\subsection{Adversarial Machine Learning}

As previously discussed, machine learning classification algorithms excel at learning functions to differentiate previously observed data, but currently fail to correctly classify data that is not similar to previously observed data. Common methods for measuring the performance of a machine learning classifier measure the \textit{generalization error} of a model by holding out a subset of the previously labeled data, the \textit{validation dataset}, to test how a model is able to generalize to data that it has not seen. This method is great for measuring how well a model generalizes to the data in the validation dataset, however it does not measure the performance of the model on data that is not represented by the validation dataset. For example, in a handwritten digit recognition data set, a small set of digits from that dataset may be withheld for testing purposes, but the withheld 4's still have the same angular shape and long straight lines as the 4's in the training set. If a data sample that has different features than any previously observed class is observed, the model has no way of correctly labelling it correctly. However, this "different" data is not represented in the validation dataset, so the measured accuracy of the model is still high.

\cite{barreno_can_2006} first discussed this issue in the context of system security. Two different types of attacks were presented against machine learning algorithms. The first of these is called an \textit{exploratory attack}, in which samples are fed to the model that allow the attacker to learn more about  the decision function of the model. This type of attack may produce a sample that is misclassified by the model, essentially exploiting the inability of a learning algorithm to adapt to unseen input data.

The second type of attack discussed is a \textit{causative attack}, which can be realized if the training data that is used to construct a model has been tampered with at all. Tampering with the training data may cause some of the salient features that are learned by a model to be modified, changing the way that it classifies future legitimate samples. This type of attack is performed at training time, as opposed to testing time exploratory attacks.

\section{Convolutional Neural Networks Under Attack}
\label{sec:convattacks}

In this section, evasion attacks are discussed in more depth, especially on CNNs. The state of the art in both attacks and defenses are then discussed, along with the problems with each.

\subsection{Evasion Attacks}

\cite{barreno_can_2006} first considered the possibility of an adversary that intelligently crafts data samples for the purpose of probing a model to discover information about the function it has learned. This concept was later explored by \cite{huang_adversarial_2011}, who introduce the idea of an \textit{Evasion Attack}, where the goal of the adversary is to craft a sample that is misclassified by the learned model. This is an exploratory attack with a more targeted goal. 

An evasion attack takes advantage of the inability of current algorithms to learn the class of data that has not been observed. Samples that are crafted using evasion attack methods are actually just samples that have the potential to exist in the environment, but live in a portion of the feature space of the environment that has not been observed frequently enough for the learning algorithm the understand how to classify it correctly.

\begin{figure}[H]
\centering
\includegraphics[width=4in,height=4in,keepaspectratio]{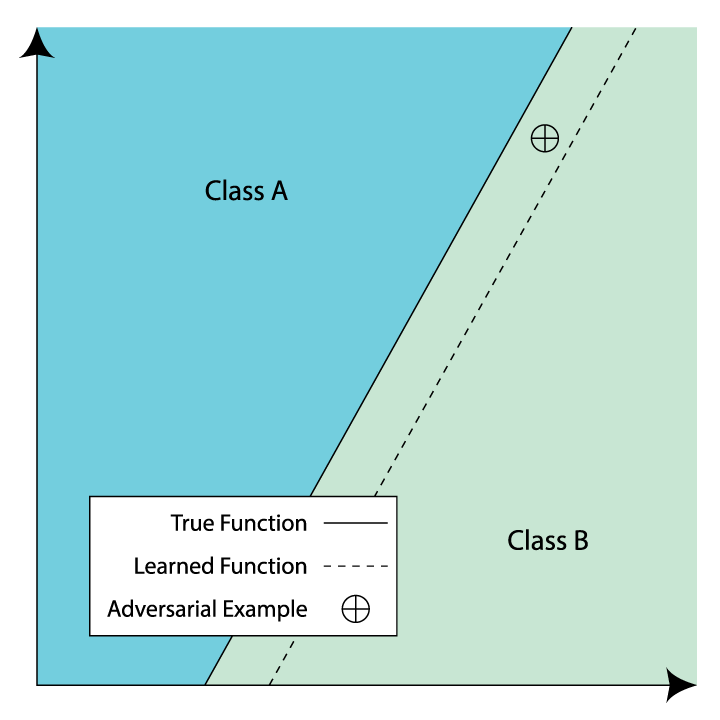}
\caption{Targeted Evasion Attack. Each axis represents a feature of the data that a model is trained on. The dashed lines represent the learned function, while the solid line represents the true function that differentiate classes A and B. The adversarial sample lies between these functions, in a section that \textit{should} be classified as class B, but is instead classified as class A.}
\label{fig:targetedAttack}
\end{figure}

In a \textit{targeted evasion attack}, the malicious sample that is send to the model is meant to be specifically misclassified as a \textit{target class}, yet truly reside in the feature space of another class. A visualization of this is provided in Figure \ref{fig:targetedAttack}.

Naturally, this vulnerability exists in Convolutional Neural Networks as well, as they are also unable to adapt to data that has not yet been observed. However, \cite{szegedy_intriguing_2013} were the first to discover how small of a perturbation to a network's input is required to cause a misclassification. They found that in the case of an image classification network, a slightly modified image can completely alter the label assigned to the modified image without causing a distinguishable difference between the two images to the human eye. When applied to a CNN, the samples produced by an evasion attack are called \textit{adversarial examples}.

\subsubsection{Adversarial Examples}

In this work, Adversarial Examples are defined as the result of a perturbation of a legitimate example such that it is classified differently than the original, legitimate sample while still truly belonging to the original class. Hence, the adversarial example is an \textit{error} made by the model on data that it does not recognize.

The true class of a sample is determined by the true unknown discriminator function that represents the data. Because this function is not known, an expert human analyst is normally used to represent the true discriminator. 

It is important to note the true class constraint that is imposed on the definition of adversarial examples. Under this definition, adversarial examples occur because of the difference between the learned function of a model and the true function of the environment. This concept is illustrated in figure \ref{fig:targetedAttack}. In the learning process, as these two functions approach each other, the potential space for adversarial examples is diminished. A true defense against adversarial examples would make this space negligible.

\textit{Distance Metrics}.
\label{sec:distances}
\cite{carlini_towards_2016} describe three different distance metrics that can be used to optimize adversarial examples; $l_{0}$ distance, $l_{2}$ distance, and $l_{\infty}$ distance. These three distances are briefly described below:

\begin{itemize}
\item \boldsymbol{$l_0$} \textbf{Distance} The number of features that have been alterred in an adversarial image.

\item \boldsymbol{$l_2$} \textbf{Distance} Calculated as the Standard Euclidean distance between source image and adversarial image, where each pixel value in an image is a coordinate.

\item \boldsymbol{$l_\infty$} \textbf{Distance} The maximum modification that is made to any feature of the source image.
\end{itemize}

\textit{Are all Adversarial Examples Dangerous?}.
As long as there is a discrepancy between the true discrimination function and a learned one, an adversarial example will always exist. However, this does not mean that all adversarial examples are equally dangerous. The success of an adversarial example is highly dependent on the  capability of the attacker and the constraints imposed by the system that is being attacked, so just because an adversarial sample exists does not mean that it is effective against the system being attacked.

For example, if an adversary is attempting to cause a self-driving car to mistake a stop sign for a "no parking" sign, they may draw on the stop sign to add a perturbation which causes it to be misclassified. If there is too much drawing, it could be visually confusing enough to the passenger in the car to invoke cautionary action such as a manual break. However, if there is only a little bit of drawing on the sign, the driver may not take notice but the attack would be successful, and the car could speed through the stop sign. Adding constraint makes the crafting of an adversarial sample more difficult. This example showcases how adversarial examples will always exist, but may not always be practical depending on the constraints imposed by the system that is being attacked.

The risk of an adversarial example is dependent on the application, model accuracy on adversarial examples, and attacker capability. A formal risk framework would be helpful in understanding the correlation between these three components, but this is out of the scope of this work.

\subsubsection{Problems Classifying with Softmax}

One common method for using neural networks as a classifier is to use the \textit{softmax} function as the final layer of the network. When performing classification with softmax, the number of neurons in the output is normally the same as the number of classes in the dataset. The softmax function squashes the network's output vector from each of these neurons to a set of values between zero and one, summing up to a value of one. Each of these values represent the class probability of each class in the dataset. 

In some specific instances, the true function of an environment should output an equal probability between multiple classes for a given input. In a neural network classification system, the standard approach to classifying an input is to use the maximum output of the network as a label. This is useful for classification purposes, but information about the output of the network is lost. For example, imagine a morphed image of an apple and an orange. In a binary classification system for an apple and an orange, this image should have a fifty percent probability of each class. This probability represents the fact that the neural network cannot suggest either class with any confidence. However, if a small change is made to the input so that it is classified as the apple class with a probability of fifty-one percent, the true discriminator function's output will be an "apple" classification, even though the image does not truly represent an apple. In fact, with this logic, the input could be modified to be classified as a true "orange" class, and the resulting image would be imperceptibly different than the "apple" image.

The previous example illustrates an inherent problem with even well trained discrimination models that can only output one class; a large number of potential samples do not truly represent any of the classes. There are two ways of dealing with this. The first would be to discard samples that are not classified with an adequately high probability in any class. The second method is to add an extra bogus class for any data that does not truly represent any other class.

\subsection{Crafting Adversarial Samples}
\label{sec:attacks}

A number of methods have been proposed for crafting adversarial examples. The goal of these methods is to find a perturbation to a real sample such that when the perturbation is applied, the sample is misclassified by a model. Different optimization procedure are employed by these attacks to find an adversarial example, depending on what the goal of the attack is. In some attacks, the goal is simply to find any perturbation that causes the original sample to be misclassified as any other class, which is considered an \textit{indiscriminate attack}. Other attacks seek to find a perturbation which causes the original sample to be misclassified a specific class, which is a \textit{targeted attack}. Furthermore, the impact that an attack can have on the original sample can also differ. Some attacks operate by changing many features by a small amount, while others modify a smaller number of features by a more substantial amount to achieve the same results.

\cite{szegedy_intriguing_2013} proposed the first method for finding adversarial examples. This method employs an optimization procedure that minimizes the loss function of an adversarial input for a target class. The method is effective, but requires multiple repetitions to find an adversarial example with a small perturbation. This iterative approach is very computationally expensive when crafting many adversarial examples.

To solve the problem of computational demand,  \cite{goodfellow_explaining_2014} proposed a method for crafting samples that maximizes the loss of a specific class, which they call the Fast Gradient Sign Method(FGSM). The optimization method in this attack is based on the sign of the gradient of a network's loss with respect to the input sample to that network, which can be calculated using backpropagation. By changing the input slightly in the direction of that sign, adversarial examples can be found quickly and with much lower computational demand. This attack is indiscriminate, so the class of the resulting image could be any other class that the model recognizes.

Two other methods have been proposed for the efficient production of adversarial samples; \cite{huang_learning_2015} propose a similar method to FGSM that follows the sign of the model output instead of the loss, and \cite{moosavi-dezfooli_deepfool:_2016} propose another method called DeepFool that also uses the gradient of the model output, but does not cause an increase in test error as FGSM does.

Other approaches have been proposed that modify less features of the original sample, but make larger modifications. \cite{papernot_limitations_2015} approach the problem by finding the \textit{forward derivative} of a network, and slowly growing the features that are modified until a core group of important features are found that can cause the misclassification. \cite{carlini_towards_2016} construct a similar targeted attack, but add a few modifications that prevent getting stuck in local extremas during the optimization process, and \textit{shrinks} the features that number of features that are modified instead of growing them. Carlini's approach leads to a more robust attack than prior attacks that finds closer adversarial examples.

\subsection{Current Defenses against Adversarial Examples}

A variety of approaches have been proposed for defending neural networks against adversarial examples. These approaches can be generally split up into two categories; sample filtration, where adversarial examples are detected by the system or model being attacked or training data augmentation, where the classification model is trained to be resilient to adversarial examples by introducing additional, modified training data. A secure system could utilize both of these methods to recognize and automatically handle malicious data, but ideally the training process should prevent "dangerous" adversarial examples from existing in the first place. 

\subsubsection{Sample Filtering}

A basic filtering defense may be to detect all samples that include data that is not possible in a specific application, such as a pixel value of 300 in an image where the maximum pixel value is 255. This anomaly detection and filtering may stop some primitive attacks, but does not prevent more advanced attacks such as modern $l_{0}$ attacks that only modify a smaller number of pixels within a legitimate range.

\textit{Defensive Distillation} was proposed by \cite{papernot_distillation} as a defense against adversarial examples. This defense is based on training a \textit{distilled network} by transferring the knowledge of a pre-trained \textit{teacher network} to another \textit{student network}, or distilled network, with fewer parameters. When training the student network, \textit{soft labels} are used that are produced by the teacher, which are advantageous in training because they include the predictive softmax output as part of the training data and contain additional information about class similarities learned by the teacher network. Classification should then be done on the distilled network. This approach claims to reduce overfitting and provide a more secure network, however \cite{carlini_defensive_2016} found that it is easily broken and does not provide robust security.

\subsubsection{Training Data Augmentation}

Current research on creating models that are robust to adversarial examples has been heavily focused on adversarial training. This process entails building adversarial examples for a model, and training on those adversarial examples with the purpose of introducing data to the model that would not otherwise be present in the environment so that the model may learn to correctly classify it. Introducing this data provides the algorithm with relevant information to better fit the true discrimination function. This idea has been briefly discussed by \cite{goodfellow_explaining_2014}, where the observation is made that the error rate of adversarial examples on a learning model continues to decrease after the error rate of the model on training samples converges to a minimum.

This approach has been effective at preventing some adversarial examples. However, the production of adversarial examples using most methods is computationally expensive so it is not practical to compute a substantial number of samples for use with this process. The Fast Gradient Sign Method(FGSM) \cite{goodfellow_explaining_2014} attempts to solve this problem by easily and quickly producing a large number of samples, but results in lower quality examples that do not adequately represent all possible types of malicious data. It is also based on the \textit{linearity hypothesis}\cite{goodfellow_explaining_2014}, and does not perform as well on models that are not represented well by this theory. Overall, adversarial training is effective at preventing any type of malicious data that can be produced, but fails to represent all possible types of adversarial samples without a substantial amount of computation.

\section{Understanding The Vulnerability of CNNs to Evasion Attacks}
\label{sec:advML}

\subsection{Linearity Hypothesis}

Currently, the most accepted theory on the existence of adversarial examples is the \textit{Linearity Hypothesis}, first explored in \cite{goodfellow_explaining_2014}. This theory hypothesizes that adversarial examples are caused by the linear behavior of a model as a function of its inputs. This inherent linear behavior is explained by the prevalence of linear building blocks for these networks, such as the commonly used rectified linear unit. To support this hypothesis, a quantitative analysis of a CNN classifier showed that the output of the network was piece-wise linear as a function of the inputs to the classifier. This hypothesis led to the creation of the Fast Gradient Sign Method(FGSM) attack.

\subsection{Adversarial Resistance}

In this work, the term \textit{adversarial resistance} is if used to describe the ability of a model to prevent "\textit{close}" adversarial examples from existing. If the closest possible adversarial example on one model is measurably farther than that of another model, the adversarial resistance of the first model is greater. This resistance can be measured by any metric that measures some type of distance. The distance could be an $l_{0}$ distance, $l_{2}$ distance, or $l_{\infty}$ distance as described in section \ref{sec:distances}, as well as any other metric that accurately represents the magnitude of change between a source sample and its corresponding adversarial example. By maximizing this resistance to close adversarial examples, a model is less prone to attacks that are successful with high confidence.

Adversarial resistance is closely related to test data generalization, but the key difference is that to have greater adversarial resistance a model must be able to correctly classify data that does not appear in the environment. In fact, it is possible for testing error of a model to reach zero on new data while the adversarial resistance of that model is low due to the existence of close adversarial examples that have not been observed. This highlights the underlying problem with neural networks that causes adversarial examples; a neural network may learn any function\cite{Hornik1989359}, but its ability to learn that function is dependent on the training data that is used to create a model.

\subsection{Considerations of the Dataset}
\label{sec:datasetDiscussion}

When considering the vulnerability a model, the similarity between individual classes in the dataset can have a drastic impact on adversarial resistance between specific source-target classes. In any dataset, one pair of classes may share more similarities than another pair of classes. For example, a classification system for animals might differentiate a very large number of animals that have a wide variety of salient features. However, some pairs of animals have similar salient features, such as a camel and a horse. A classification system may be able to differentiate the two, but a smaller perturbation may be required to misclassify a horse as a camel than a horse as another animal, such as a lizard.

This imbalance in the similarity between classes in a model poses an added challenge to defending a classification system; similarly featured classes are likely able to be perturbed to be misclassified as each other more easily than classes with less similar features. This is inherent to the functionality of machine learning classifiers because they seek to learn the true representative function of their environment, which in the case of image classification of horses and camels is a function that has two different outputs for very similar inputs.

Preventing this imbalance from existing would involve training the classifier to learn that camels and horses \textit{do not} look similar, which is not desirable and undermines the utility of a classification system. Instead, the most effective way to combat this issue is to understand what these similar classes in a dataset are, and find ways to extract features that better represent the subtle differences in similar images.

\subsection{Measuring Class Distances}

It is easy for a human eye to visually distinguish classes and understand how much two classes differ, but a modern machine learning classifier gives no insight into this difference between the classes it has trained on. In some ways, this is the goal of machine learning classification problems; find a function that separates a number of classes in the most optimal matter, which may be defined by a distance metric of some type. However, this distance cannot be directly extracted from a trained model, as the only output of a model is the class prediction. In order to extract the model's understanding of distance between classes, more creative approaches must be used.

\cite{papernot_thesis} first quantified the \textit{adversarial distance} of an adversarial example from its original source sample, providing insight into the distance between classes in a dataset. This distance was calculated using the results of the first iteration of the JSMA attack presented in that paper, which was found to be a good predictor of the average distortion required to misclassify each sample relative to its success rate, which is referred to as \textit{hardness}.

In Papernot's work, the adversarial distance metric is used as a predictive measurement of adversarial distance, while hardness is used as an absolute, quantifiable distance. In this work, this definition is changed so that adversarial distance represents the true distance between a class pair. To quantify this distance, a targeted attack is performed on a sample belonging to one class to find the "closest" sample in another. Instead of using Papernot's $l_{0}$ attack, the $l_{2}$ attack described in \cite{carlini_towards_2016} and discussed in section \ref{sec:attacks} is used instead. This attack is optimized over a distance metric instead of the number of modified pixels that is optimized over in an $l_{0}$ attack, providing a more realistic view of the closest sample in a target class. The resulting metric is more closely related to hardness in the previous work.

The distance metric that is optimized in Carlini's approach is the standard euclidean distance of a source sample and adversarial example, defined as:

\begin{equation}
\label{eq:euclidean}
D(X^{*}, X) = \big( \sum_{i=1}^{n} |X^{*}_{i} - X_{i}|^2 \big) ^ \frac{1}{2}
\end{equation}

where $X^{*}$ is the adversarial example, $X$ is the source sample, and $i$ is each feature in the samples. By measuring this distance for each class pair, a deeper understanding of the distance between classes learned by the classifier can be provided. 

The naive approach to collecting this metric would be to take any sample from each source class and measure the distance to each target class using the methodology described above. The problem with this approach is that each sample in a class is different, and some samples may contain features that position them more closely to a target class than other samples in the data set. However, there is no true representative image of each source class, so a best estimate of each must be found.

\begin{figure}[H]
\centering
\includegraphics[width=\textwidth,height=\textheight,keepaspectratio]{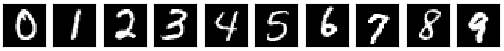}
\caption{Images found to best represent classes learned by LeNet Model.}
\label{fig:bestEst}
\end{figure}

\textit{Measuring MNIST Similarities}
\label{sec:mnistSim}
MNIST is a good data set to experiment on to find adversarial distances due to the ease of training a classification model for MNIST and the presence of varying degrees of similarity between classes. For instance, "3" is far more similar to an "8" than a "4", and the distances found should reflect this.

To find images in MNIST that most represent the class they belong to and therefore make good candidates for experimentation, the average adversarial distance for each target class is found for 100 samples in each class. This average distance is compared to the adversarial distance of each source sample, and the sample with the smallest difference in adversarial distance to the average is chosen as the best estimate sample of each class. These images are shown in figure \ref{fig:bestEst}.

\begin{figure}
\centering
\includegraphics[width=3.5in,height=3.5in,keepaspectratio]{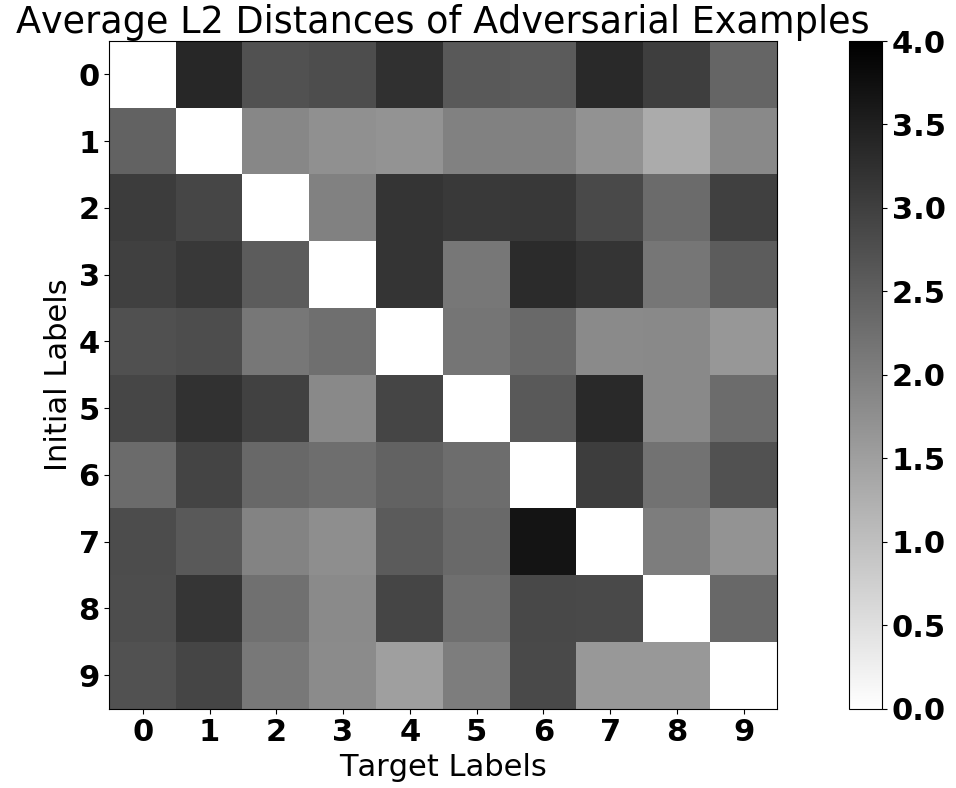}
\caption{Average adversarial distance of 100 source and target class pairs. Each row represents a source class, and each column represents a target class.}
\label{fig:mnistEval}
\end{figure}

\begin{figure}
\centering
\includegraphics[width=3.5in,height=3.5in,keepaspectratio]{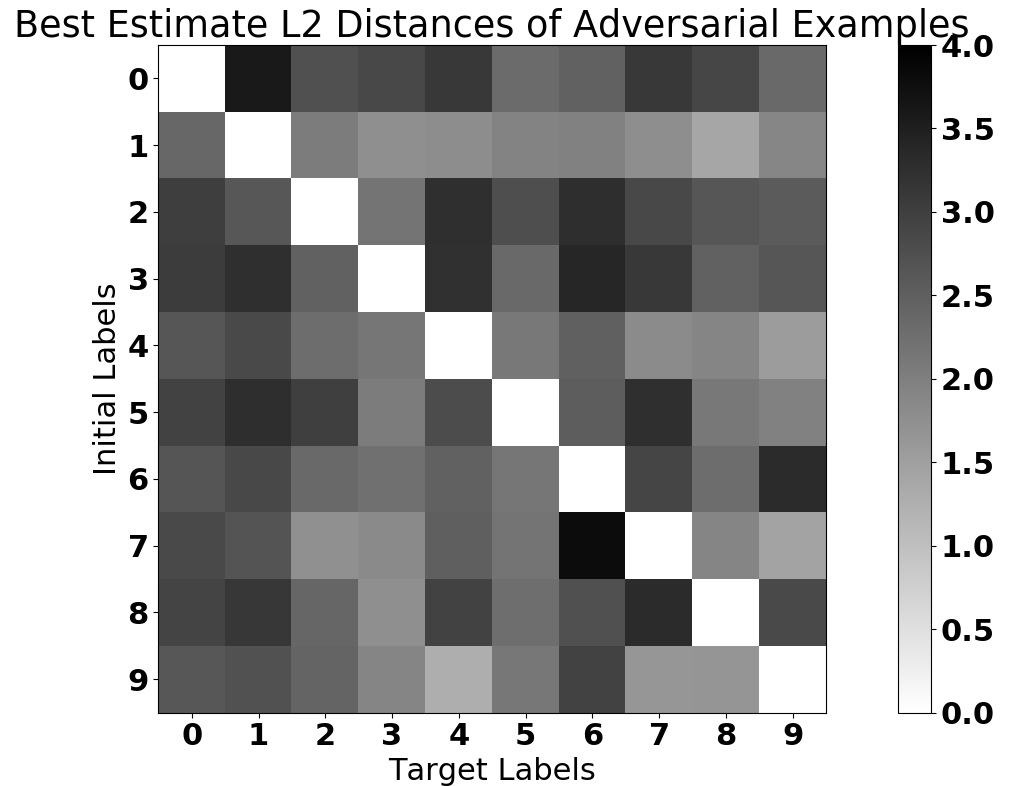}
\caption{Adversarial distance of source and target class pairs, as measured with best-estimate class samples. Each row represents a source class, and each column represents a target class.}
\label{fig:bestEstEval}
\end{figure}

The average adversarial distance values for a large number of samples in each class is a good representation of the true distances between classes in MNIST, however the production of these adversarial examples is very expensive using the targeted $l_{2}$ attack, and is not computationally feasible for extensive experimentation. Instead, the best estimate samples can be used instead to represent their respective classes and greatly reduce the amount of computation required to run experiments with the attack used in this paper. Figure \ref{fig:mnistEval} shows the adversarial distances found using the average values of each source class, while figure \ref{fig:bestEstEval} shows the adversarial distances found using the best estimate classes. The resulting distances between classes and patterns are comparable, and show that the best estimate classes extracted are good estimators of the true adversarial distance between a source and target class.

\section{Methods}
\label{sec:methods}

In this section, the details of the experiments conducting in this work are described. This includes the experimental design, the attack methodology, the dataset used, the model creation process, and the software used to conduct the experiments. The reasoning behind each of the choices made is also described in this chapter.

\subsection{Architecture and Models}

The LeNet architecture is used as the control for experiments in this work. This architecture has been well studied in the past, so experimenting on this basic architecture allows easy comparison of the results in this work to other experiments on LeNet. Because it is the basis of many modern architectures such as GoogLeNet, experimentation on LeNet should provide insight into how modern systems behave under the circumstances tested in this work.

The original LeNet model used for comparison consists of two convolution layers with max-pooling after each, followed by a single dense layer. A standard gradient descent optimizer with momentum is used to train the model until it reaches \textbf{at least 98\% accuracy}.

\begin{table}[]
\centering
\caption{Hyperparameters used to train basic LeNet Model}
\label{tab:controlModel}
\begin{tabular}{|c|c|}
\hline
\textbf{Parameter} & \textbf{Value} \\
\hline

Activation Unit & ReLU  \\
\hline

Dropout on Input Layer& 0\% \\
\hline

Dropout on Dense Layer& 0\% \\
\hline

\makecell{Convolution Kernel Size\\(Dimensions)} & (5,5) \\
\hline

\makecell{Max Pooling Size\\(Dimensions)} & (2,2)\\
\hline

Weight Initializers & Glorot Uniform\\
\hline

Bias Initializers & Zeros \\
\hline

Number of Hidden Units & 500\\
\hline

Number of Dense Layers & 1\\
\hline

\makecell{Number of Filters\\per CNN} & 32 First Layer, 64 Second\\
\hline

Kernel Regularizers & None \\
\hline

Acivity Regularizers & None\\
\hline

\makecell{Stride Size} & (1,1)\\
\hline

Optimizer & SGD\\
\hline

Momentum & 0.9\\
\hline

Loss & Cross Entropy\\
\hline

\end{tabular}
\end{table}

The hyperparameters used to train the control model are described in table \ref{tab:controlModel}. Every other model that is produced is created in the same way as the original LeNet model, but with a variation of one of the hyperparameters of either the model itself or the training process.

\subsection{Attack Algorithm}
\label{sec:attack}

A variety of methods have been developed for crafting adversarial examples, as previously discussed in section \ref{sec:attacks}. Of these methods, the approach in \cite{carlini_towards_2016} is chosen to evaluate the models in this work. The reason is two-fold; first, this attack has a 100\% success rate of crafting adversarial examples on the data set used in this work. This is desirable because it allows for a consistent comparison of various models without concern about the attack's ability to find an adversarial sample. Second, this attack is targeted, and contains a parameter that controls the confidence with which the model will classify the adversarial example as the target class. This parameter is useful when measuring the best and worst case adversarial examples on a model, because it helps control the confidence of prediction. By setting this parameter to 0, it is also possible to find the shortest $l_{2}$ distance between a source sample and target class with a targeted attack, which is helpful for understanding the vulnerability between classes.

The targeted version of this attack is selected for use because it allows more extensive comparison of inter-class adversarial distances between arbitrary source-target classes. Without this ability, the importance of the resemblance between two labels(such as a '3' and an '8') is lost, and there is no way to consider the impact that the classes in a dataset have on the model's adversarial resistance. With a targeted attack, the relationship between pairs of labels can be more closely examined, and the possibility of vulnerability that is inherent to the data set can be better understood.

\subsubsection{Attack Details}
\label{sec:attackCompare}

Due to computational limitations, extensive testing of a large number of each source-target class pair for each model that is constructed is not feasible. Instead, the best-estimate samples found in section \ref{sec:mnistSim} are used to compute an estimated value for each source-target class pair. It should be noted that the function learned on this model is not perfect and therefore the adversarial distances for these classes is likely not correct; however, it provides this work with a point of reference for the analysis of further models.

\subsection{Experimental Design}

Due to the large number of hyperparameters that are available when training a convolutional neural network, as well as the time required to train each model, training every possible combination of hyperparameters was not feasible for this work. Instead, a control model was built to compare other models against, which is the LeNet model previously described, and hyperparameters were changed one at a time for each new model. The attack discussed in section \ref{sec:attack} is then implemented on each model that is produced using this methodology. Each of these models is built on the same set of training data. 

\subsubsection{Dataset}

The MNIST dataset is used for all experiments in this work. This is a well explored dataset in the machine learning community, so it should make this work easy to compare to other works. Furthermore, it is a somewhat small dataset with significantly different classes, which makes it manageable for the required experiments and also useful for making attack comparisons.

The MNIST training dataset consists of 60,000 images, each represented as a vector of 784 features. Each feature represents a pixel intensity beween the values of 0 and 255. These values are normalized to a value between -.5 and .5, and used as training data for all of the models in this work.

Making meaningful measurements for comparing attacks requires an understanding of the inherent similarities between classes in MNIST. If two classes have similar salient features, a targeted attack between these two classes should not be compared to a targeted attack between two less similar classes.

In an ideal model, these similarities and differences correspond to the similarities and differences that humans observe between different digits, which should be reflected in the measurements gathered during these attacks. However, this is not guaranteed behavior, so before conducting attack comparisons on various models, an experiment is run to observe which classes in MNIST are similar. Similarity is determined by how much modification is required to cause one class to be misclassified as another using the crafting approach proposed by \cite{carlini_towards_2016}.

\subsubsection{Software Used}

These experiments use code adapted from the work of \cite{carlini_towards_2016} to ensure consistency in the attack, as well as the data used to train the model. The network architecture itself is modified from this work to more closely resemble the original LeNet architecture. The architecture used in Carlini's paper includes four convolution layers, with a max-pooling layer after the second and fourth convolution. Two dense layers were used with 200 neurons each, and a 50\% dropout in between layers. In the modified version for this work, there are only two convolution layers with max-pooling after each, and a single dense layer with no dropout. This code makes use of the Tensorflow \cite{tensorflow} and Keras \cite{keras} libraries to run these experiments. For all data analysis, the NumPy \cite{numpy} and Matplotlib \cite{Hunter:2007} libraries are used.

\subsection{Measurement Methodology}
\label{sec:measurement}

Previous works have examined the success of attacks that craft adversarial samples by measuring the percentage of adversarial examples that were classified incorrectly by a model but were classified correctly before the perturbation was applied. This measurement methodology depends on the fact that an algorithm may not find a perturbation to an input sample that causes it to become misclassified. Given a powerful enough attack, this assumption is invalid because it is always possible to create an adversarial sample when the learned function does not perfectly represent the true function of its environment, and therefore this metric is not useful because a reliable attack will always yield 100\% attack accuracy. Furthermore, even if a learned function is accurate and there is little room for a true adversarial example to exist, there may still be dangerous examples that do not adequately represent any class but still must be classified. As future optimized attacks are created which always find adversarial samples(or samples that are difficult to label), the success and impact of adversarial samples must be measured through different means.

In this work, Carlini's and Wagner's approach for crafting targeted adversarial samples in order to compare the ability of a model to resist evasion attacks. The case is considered in which an image must be crafted that is visually similar to the original sample, but is classified as the target class instead.  In the following section, the metrics used to conduct the experiments in this work are discussed.

\subsubsection{Comparing Results}
\label{sec:compareRes}
With a 100\% success rate of crafting adversarial examples, other metrics must be used to understand how much adversarial resistance the model has. This can be done by measuring how much modification must be made to a legitimate example in order to cause the misclassification. In this work, \textit{adversarial distance} is measured using the $l_{2}$ distance between a legitimate sample and the corresponding adversarial example that is built from it. 

$l_{2}$ distance refers to the standard Euclidean distance between each sample. This metric is an indicator of how difficult it is for Carlini's attack to succeed, as this is the metric that is optimized over in the attack itself. A higher $l_{2}$ distance between a source and adversarial example signifies that the attack was more difficult and less effective. This metric also generalizes well to any type of dataset because it is solely reliant on the numerical values of the data itself with no application specific dependence.

$l_{2}$ distance is not the only possible measurement for adversarial distance.  \cite{papernot_thesis} explored \textit{distortion} of an image as an adversarial distance metric in the context of an $l_{0}$ attack. The metric used in that work measures the percentage of pixels that are modified, regardless of the percent of modification that is made. This is a relevant metric on $l_{0}$ attacks, because in this type of attack, the goal is to modify the smallest number of pixels possible while modifying individual pixels to a greater extent, so the assumption in using percentage of modified pixels as a metric is that every pixel that is modified is visually significant enough to impact a human's perception of the image. For an $l_{2}$ attack such as the one explored in this work, this metric is not useful because the percent of distortion in each pixel in an image is much lower and sometimes unnoticeable.

\begin{figure}[H]
\centering
\includegraphics[width=\textwidth,height=\textheight,keepaspectratio]{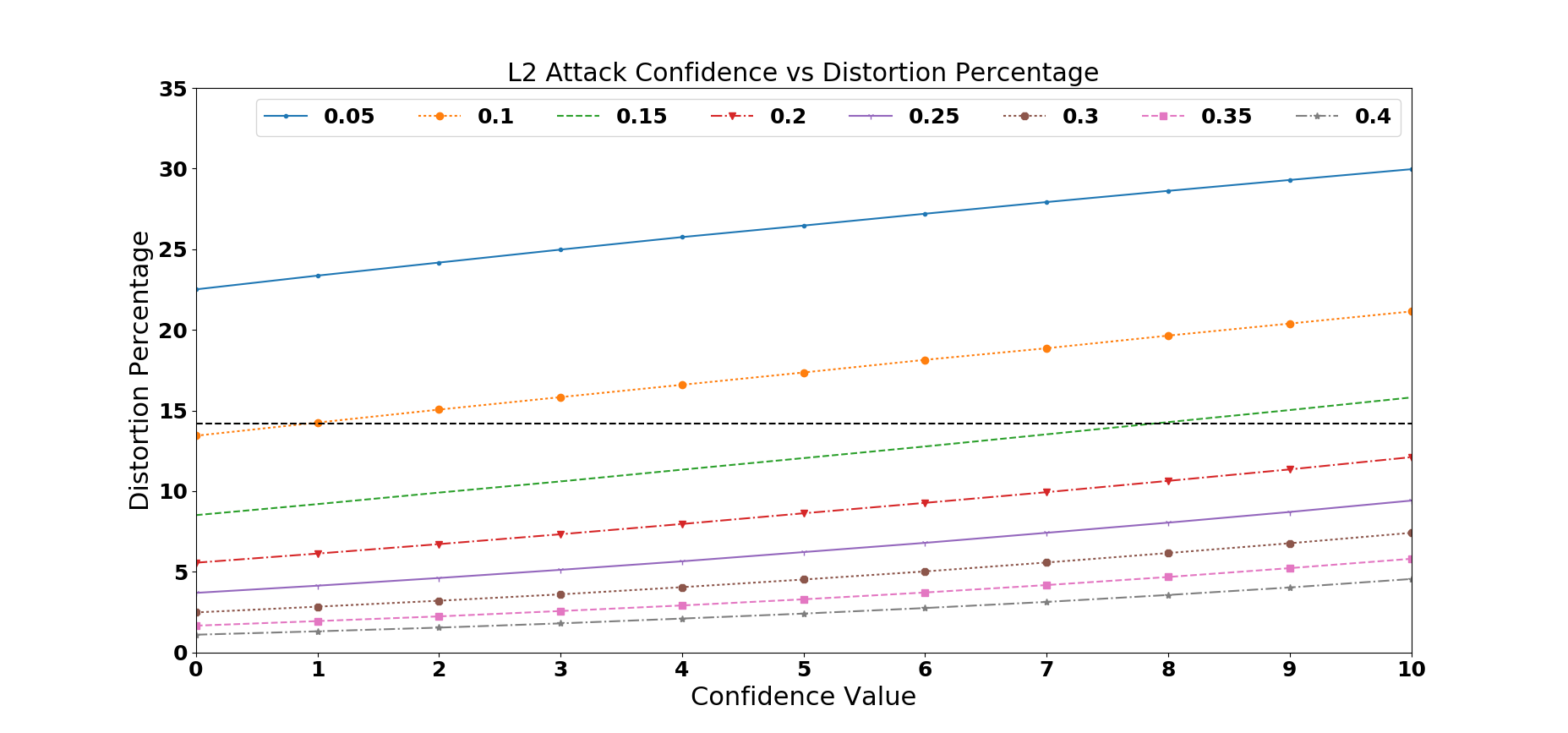}
\caption{Average Distortion of a successful attack for confidence values ranging from 0-10. Each line represents the average percent distortion of the input images when measured with a different minimum distortion constraint. If the minimum distortion is set to 0, the graph is simply a flat line at 100\% distortion, because the $l_{2}$ attack modifies every single pixel in the image by at least a small amount. The dashed line at 14.2\% represents the threshold that humans will likely misclassify adversarial images.}
\label{fig:Distortion}
\end{figure}

Instead, there are two distortion metrics that can be helpful in understanding adversarial distance for an $l_{2}$ attack. The first is to use percentage of pixels that have been modified in an adversarial image, with an added constraint that only pixels that have been modified over a certain value can contribute to this value. This constraint adds the ability to represent only pixel modifications that are noticeable by a human. For instance, if a pixel is modified by only 1\%, it may not be noticeable, and the measurement would not factor this image in. However, if a pixel is modified by 35\%, it is likely more noticeable visually, so the metric accounts for this image in the distortion value. Figure \ref{fig:Distortion} shows the impact that the confidence parameter in the $l_{2}$ attack used in this work impacts the distortion of an adversarial example on the default LeNet model, with the minimum distortion parameters for this metric ranging from 5\% to 40\% distortion.

Instead of measuring the percentage of modified pixels, the second potential distortion metric is the average percent of distortion per pixel in each image. This metric is a cumulative total of the amount of distortion that is applied to an image. While this metric more accurately represents the true distortion applied to an image, it behaves in the exact same way as the $l_{2}$ metric, and provides no addition useful information about the distance between a source sample and an adversarial example.

Similarly to $l_{2}$ distance, a larger value for this metric represents a more secure system because it means a greater change to the input data is required to find an adversarial example. This is also a more intuitive measurement visually, is it represents the magnitude of change visually seen in the picture more directly.

A higher adversarial distance means a larger change in the sample is required for a successful attack, so there is greater adversarial resistance. A smaller adversarial distance means only a small modification is required to find an adversarial example. Visually, a greater adversarial distance causes a greater change in the pixel values of an image, causing the image to represent the original class less.

To compare each model in this work, the average adversarial distance of all source-target class pairs on a model is measured, as well as the maximum distance, minimum distance, and standard deviation for each model. The standard deviation for each model represents the variance in source-target distances for a model; the smaller this value is, the more similarly distanced each class pair is.

\subsection{Quantifying Parameter impact}
\label{sec:quantifyResults}

In order to determine the statistical significance of the mean, minimum, maximum, and variation values for each tested model, a statistical method must be used to quantify \textit{how} different an observation on a modified model is in comparison with the normal range of potential observations on the base LeNet model.

To quantify this measurement, 100 different LeNet models were created, and each source-target class pair was attacked using the same methodology that is used in the rest of this section for each model. The source samples used are also the same best-estimate samples found in section \ref{sec:mnistSim}, which are used in all other attacks in this section as well.

\begin{table}[H]
\caption{Standard deviation of distance measurements across 100 LeNet models. Note how small these values are; most models that were trained had very similar characteristics.}
\label{tab:lenetAnalysis}
\centering
    \begin{tabular}{|c|c|c|c|}
    \hline
    Mean Distance & Min Distance & Max Distance & Std Deviation\\
    \hline
    0.03 & 0.1 & 0.09 & 0.02\\
    \hline
\end{tabular}
\end{table}

The mean, minimum, maximum, and standard deviation distance measurements across all source-target pairs for each of the 100 trained models is recorded, and averaged out to find the values that will be used for comparison to experimental models in this section. For each of the four measurements, the standard deviation of the measurement across the 100 samples is taken and used to quantify how strange an observation is with respect to the original LeNet model. These standard deviations are listed in table \ref{tab:lenetAnalysis}.

For each model that is tested in the following sections, the number of standard deviations away from the LeNet average for each metric is used to determine how significant of an impact a parameter has on a model. If the variance between the specific parameters tested in each experiment is between 3-5 standard deviations of the normal for these metrics, the impact is low. 5-7 standard deviations is a medium impact, and greater than 7 standard deviations from the normal is a high impact parameter.

\section{Results}
\label{sec:caseStudy}
Using the previously described methodology, the results of those experiments are presented and discussed in this section. The findings from these experiments shows the impact of hyperparameters and training methods on the vulnerability of CNNs. First, the impact of specific network hyperparameters on adversarial distance is presented. Following this, the high impact network hyperparameters are experimented on in more detail, and those results are discussed in depth. A similar analysis is also done on a variety of training hyperparameters.  Finally, a combination of the high impact hyperparameters found is explored, and the resulting model is analyzed to determine if it provides additional security against evasion attacks.

\subsection{Network Hyperparameters}
In this section, the impact of a set of network hyperparameters on the adversarial resistance of a model is compared. Most of the tested hyperparameters do indeed have at least a small statistically significant impact on the adversarial resistance of a model. This means that when using the analysis method described above, the impact of each parameter on the mean adversarial distance of all source-target pairs was outside of three standard deviations from the expected value recorded on the original LeNet model. 

\begin{table}
\centering
\caption{Impact of network hyperparameters on adversarial distance. See table \ref{fig:allspecificParams} in the appendix for specific parameters tested.}
\label{fig:modelParams}
\resizebox{\columnwidth}{!}{
\begin{tabular}{|c|c|c|c|}
\hline
Parameter  & \multicolumn{3}{c|}{Ceteris Paribus, Impact on Perturbation Magnitude} \\

\cline{2-4}

& Impact & Impact & Brief Description of Impact \\
\hline

Activation Unit & Y & High & \makecell{Varying impacts} \\
\hline

\makecell{Percent Dropout\\on Input Layer}& Y & High & \makecell{High percents of dropout increase mean\\adversarial distance and decrease variance} \\
\hline

\makecell{Percent Dropout\\on Dense Layer}& Y & Low & \makecell{Small percentages of dropout cause\\ an increase in adversarial distance} \\
\hline

\makecell{Convolution Kernel Size\\(Dimensions)} & Y & Medium &\makecell{Smaller kernel sizes cause\\an increase in adversarial distance } \\
\hline

\makecell{Max Pooling Filter Size\\(Dimensions)} & Y & High &\makecell{Larger filters cause\\ significantly smaller adversarial distances} \\
\hline

Weight Initializers & N & None & \makecell{No significant impact\\on adversarial distance observed} \\
\hline

Bias Initializers & Y & Low  &\makecell{Random Initialization causes\\a slight increase in adversarial distance} \\
\hline

Number of Hidden Units & Y & Low & \makecell{Adversarial distance is greater when model\\ is trained with 500 units or 200 units} \\
\hline

Number of Dense Layers & N & None & \makecell{No significant impact\\on adversarial distance observed} \\
\hline

\makecell{Number of Filters\\per CNN} & Y & Low &\makecell{Increasing the number of filters slightly\\ increases mean adversarial distance} \\
\hline

Kernel Regularizers & N & None &\makecell{No significant impact\\on adversarial distance observed.} \\
\hline

Acivity Regularizers & X & X &\makecell{Could not achieve high accuracy} \\
\hline

\makecell{Stride Size} & Y & High &\makecell{Stride of size 1 causes\\lower adversarial distance} \\
\hline

\end{tabular}
}
\end{table}

In table \ref{fig:modelParams}, the impact of each hyperparameter is labeled as high, medium, or low, based on the difference in average adversarial distance for all source-target pairs. A few of these parameters, including weight initializers, number of dense layers, and kernel regularization, did not have a statistically significant impact on the adversarial distance between classes. Activity regularization was also not measurable, because a model could not be trained that had an accuracy over 98\% while also containing significant values for any activity regularization. Below 98\% accuracy, a model is considered unusable for comparison in this work, because it is unable to learn a good discriminatory function.

The low impact hyperparameters in table \ref{fig:modelParams} have a significant enough impact to warrant mention, but do not contribute any practical enhancement to the adversarial resistance of a model. The parameters that fall in this category, such as number of filters per convolution or number of hidden units, are generally modified to increase the model's ability to fit complicated data, so it is goes with intuition that they do not provide any significant ability to generalize to data that has not been observed, such as an adversarial example.

On the other hand, the high and medium impact network hyperparameters more generally impact the shape of the data samples and the feature extraction process. The statistical impact of these parameters on the adversarial distance of source-target classes in a model is much greater than that of the low impact parameters, but the actual practical enhancement to the security of these systems may not be as promising. These parameters are further discussed in section \ref{sec:highImpact}.

\subsection{Training Hyperparameters}

In this section, the impact of training hyperparameters on the adversarial resistance of a model is discussed. The same methodology is used for this comparison as was used for the comparison of network hyperparameters; the original LeNet model is trained once for each test, and each test a training hyperparameters is modified.

Intuition would say that training hyperparameters should have no impact on the adversarial resistance of a model, because these parameters should only be used to converge more quickly on an optimal solution. However, the only parameter tested that yielded no statistically significant impact was momentum. Furthermore, the choice of optimizer seemed to have as much impact on the resulting model's adversarial resistance as some network hyperparameters, such as max pool size. This comes as a surprise, as the optimizer used to train a network has no impact on the feature extraction process or the shape of the observed data. Table \ref{fig:trainingParams} show the level of impact that each of the tested training hyperparameters have according to the methodology discussed in section \ref{sec:quantifyResults}.

\begin{table}
\centering
\caption{Impact of training hyperparameters on adversarial distance.}
\label{fig:trainingParams}
\begin{tabular}{|c|c|c|c|}
\hline
Parameter  & \multicolumn{3}{c|}{Ceteris Paribus, Impact on Adversarial Distance} \\

\cline{2-4}

& Impact & Impact & Brief Description of Impact \\
\hline

Batch Size & Y & Low & \makecell{Larger batch size cause a slight\\ increase in variance, and a lower average distance} \\
\hline

Loss& Y & Low & \makecell{Adversarial distance decreased on\\models trained with Hinge Loss and MSE } \\
\hline

Optimizers & Y & High & \makecell{Varying Impacts, see table \ref{tab:highTraining}}\\
\hline

Momentum & N & None &\makecell{No significant impact\\on adversarial distance observed}\\
\hline

\end{tabular}
\end{table}

\subsection{High-Impact Parameters}
\label{sec:highImpact}

A small subsection of the hyperparameters explored in this work had a greater impact on the adversarial resistance of a model. To compare the impact that these hyperparameters had on their resulting model, the average adversarial distance of each class pair is found. The average, minimum, maximum, and standard deviation of adversarial distances for all of these class pairs is compared in tables \ref{tab:highImpactModel} and \ref{tab:highTraining}. These samples are all produced with an attack confidence value of 0, which produces a sample that is barely classified as the target sample so that the shortest distance from each source class to each target class is discovered.

\begin{table}
\centering
\caption{Comparison of medium and high impact network hyper-parameters on adversarial distance.}
\label{tab:highImpactModel}
\begin{tabular}{|l|l|l|l|l|l|}
\hline
Parameter & Specific  & \multicolumn{4}{c|}{Ceteris Paribus, Impact of Parameter Perturbation} \\
\cline{3-6}
& & Mean Distance & Min Distance & Max Distance & Std Deviation\\

\hline
LeNet Model & NA & 2.28 & 1.21  & 3.84 & 0.93 \\
\hline
Activation Unit & ReLU & 2.25 & 1.34 & 3.71 & 0.91 \\
\cline{2-6}
&  TanH  & 2.14  & 0.86  & 3.68 & 0.87 \\
\cline{2-6}
&  Sigmoid  & 2.26 & 1.31 & 3.96 & 0.98\\
\cline{2-6}
&  Linear  & 1.91  & 1.07 &   3.33 & 0.87\\
\cline{2-6}
&  ELU  & 2.18 & 1.20 &  3.50  & 0.88\\
\hline
\makecell{Percent Dropout\\on Input Layer} & 10\% & 2.37 & 1.36 & 3.71  & 0.92 \\
\cline{2-6}
& 20\% & 2.43 & 1.12 & 3.75 & 0.91 \\
\cline{2-6}
&  30\%  & 2.51  & 1.26 &  3.90  & 0.97 \\
\cline{2-6} 
&  40\%  & 2.57 & 1.53 &  4.12  & 1.00 \\
\cline{2-6}
&  50\%  & 2.63 & 1.60 &  4.02  & 1.02 \\
\hline
Kernel Size & (2, 2) & 2.05 & 0.96  & 3.28 & 0.82 \\
\cline{2-6}
&  (3, 3)  & 2.17  & 0.95  & 3.66 &0.91\\
\cline{2-6}
&  (4, 4)  & 2.21 & 1.07 &  3.68  &0.92\\
\cline{2-6}
&  (5, 5)  & 2.23 & 1.18 &  3.90  &0.92\\
\cline{2-6}
&  (6, 6)  & 2.26  & 1.00 &  3.75  &0.93\\
\hline
Pool Filter Size & (2, 2) & 2.29 & 1.09  & 3.70 & 0.94 \\
\cline{2-6}
&  (3, 3)  & 2.18  & 1.39  & 3.67 &  0.86\\
\cline{2-6}
&  (4, 4)  & 1.92 & 0.81 & 3.31 & 0.75 \\
\cline{2-6}
&  (5, 5)  & 1.89 & 0.82 &  3.10  & 0.80\\
\cline{2-6}
&  (6, 6)  & 1.84 & 1.06  & 3.12 & 0.77 \\
\hline
Stride Size & (2, 2) & 1.60 & 0.87  & 2.79 & 0.65\\
\hline

\end{tabular}
\end{table}

\begin{table}
\centering
\caption{Comparing optimizer impact on adversarial distance.}
\label{tab:highTraining}
\begin{tabular}{|l|l|l|l|l|l|}
\hline
Parameter & Specific  & \multicolumn{4}{c|}{Ceteris Paribus, Impact of Parameter on Adversarial Distance} \\
\cline{3-6}
& & Mean Distance & Min Distance & Max Distance & Std Deviation\\

\hline
LeNet Model & NA & 2.28 & 1.21  & 3.84 & 0.93 \\
\hline
Optimizers & AdaDelta & 2.27 & 1.31 & 3.75 & 0.93\\
\cline{2-6}
&  AdaGrad  & 2.12  & 0.96  & 3.96 &  0.89\\
\cline{2-6}
&  Adam  & 2.16  & 0.82  & 3.58 &  0.89\\
\cline{2-6}
&  Adamax  & 2.19  & 1.05  & 3.56 &  0.90\\
\cline{2-6}
&  Nadam  & 2.04  & 1.01  & 3.14 &  0.80\\
\cline{2-6}
&  RMS  & 1.98  & 0.70  & 3.36 &  0.86\\
\hline

\end{tabular}
\end{table}

\textit{Activation Units}. 
The activation unit that is used in the network is largely unimportant other than the impact of the tanh and linear activation functions. The ReLU, ELU, and sigmoid functions all behave similarly except for the somewhat larger maximum adversarial distance found when the network is trained with a sigmoid activation function. When a tanh function is used, the opposite occurs, and the minimum adversarial distance between classes dropped to a value of 0.86 as opposed to the highest minimum value observed at 1.34. The linear activation function yields the most dramatically different distance results, with a mean value of 1.91 as opposed to the highest observed at 2.28, and a maximum value of only 3.33 as opposed to the high max value of 3.96 when a sigmoid activation function is used.

These results are not surprising, as Linear activation functions are not generally recommended for better generalization, or better performance in general. The ReLU activation function and the Sigmoid activation function were found to contribute the most to adversarial resistance.

\textit{Percent Dropout on Input Layer}.
Although dropout on the dense layer did not yield promising results, dropout on the input layer of the network led to much higher adversarial distances. As the percentage dropout in the input layer increases, the mean distance, min distance, and max distance of the adversarial examples in the measured models also increased significantly. At 50\% dropout, the mean distance was found to be 2.63, as opposed to the average distance of 2.28 found on the base LeNet model. The maximum observed is found to be 4.02, and the minimum is 1.60. All three of these measurements are the highest recorded in all of the experiments in this work, save for the maximum distance of 4.12 observed at a dropout level of 40\%.

\textit{Kernel Size}.
Lower convolution kernel sizes produced lower adversarial distances, as well as less variation in adversarial distances between different source-target classes. With a convolution kernel size of $(2,2)$, the mean adversarial distance dropped down to 2.05, the minimum dropped to 0.95, the maximum became 3.28, and the standard deviation between source-target class distances dropped to 0.82. These values all see an upward trend as the kernel size increases. 

\textit{Pool Size}.
Lower max pooling sizes yield greater adversarial distances, and hence more secure models. A pool size of $(6,6)$ has some of the worst results, with a mean distance of 1.84 and a maximum of 3.12. As pool sizes increase, the variance in adversarial distance between classes decreases, so it requires much smaller amounts of distortion to misclassify any target from any source sample. Max pool size causes the most significant decrease in adversarial distance of any of the metrics observed in this work, save stride.

\textit{Stride}. 
Greater strides resulted in much lower adversarial distance between classes. A stride of $(2,2)$ caused the trained model to learn classes that were very close together, with a maximum distance of only 2.79 and a mean distance of 1.60, the lowest seen in this work. 

\textit{Optimizer}.
The choice of an optimizer has a larger impact on average adversarial distance than one might expect. Standard stochastic gradient descent(SGD) with momentum yields a model with the best average adversarial distance among the models trained with varying optimizers, although AdaDelta \cite{adadelta} performed just as well. Other commonly used optimizers such as Adam \cite{adam} and AdaGrad \cite{adagrad} created models with lower adversarial resistance, although they managed to converge at a high accuracy level on the validation dataset more quickly than SGD. The root mean squared(RMS) optimizer crafted the worst model, which had the lowest minimum distance found in the experiments conducted in this work.

\subsection{Building a More Secure Model}

By observing table \ref{tab:highImpactModel} it is clear that certain choices in hyperparameters of a network can have a large impact on the adversarial resistance of the resulting model. It seems that standard choices for values such as pool size, kernel size, stride, and activation function choice, such as those chosen for the standard LeNet model in this work, can help a network generalize both to valid data as well as adversarial examples. Furthermore, dropout is also observed to have a high impact on the adversarial distance of the samples produced in these experiments in comparison with other hyperparameters tested in this work. The question is, \textit{is there a combination of these hyperparameters that maximally increases a model's adversarial resistance?}

To test if this is true, a model is built that includes the values of each hyperparameter that maximally increased adversarial distance of malicious samples in the previous experiments. This model includes dropout of 40\% on the input layer, dropout of 10\% on the dense layer, a large convolution kernel size of (6,6), a stride size of (1,1), small max pooling sizes of (2,2), and the ReLU activation function. The convolution layer kernels are initialized with random values, as this was identified to have a low impact as well.

The resulting model performed similarly to the model that only included the addition of dropout on the input layer. The mean adversarial distance measured was 2.58, which is not significantly different than the mean distance measured on models trained with dropout of 40\% and 50\%.

\begin{figure}
\centering
\includegraphics[width=\textwidth,height=\textheight,keepaspectratio]{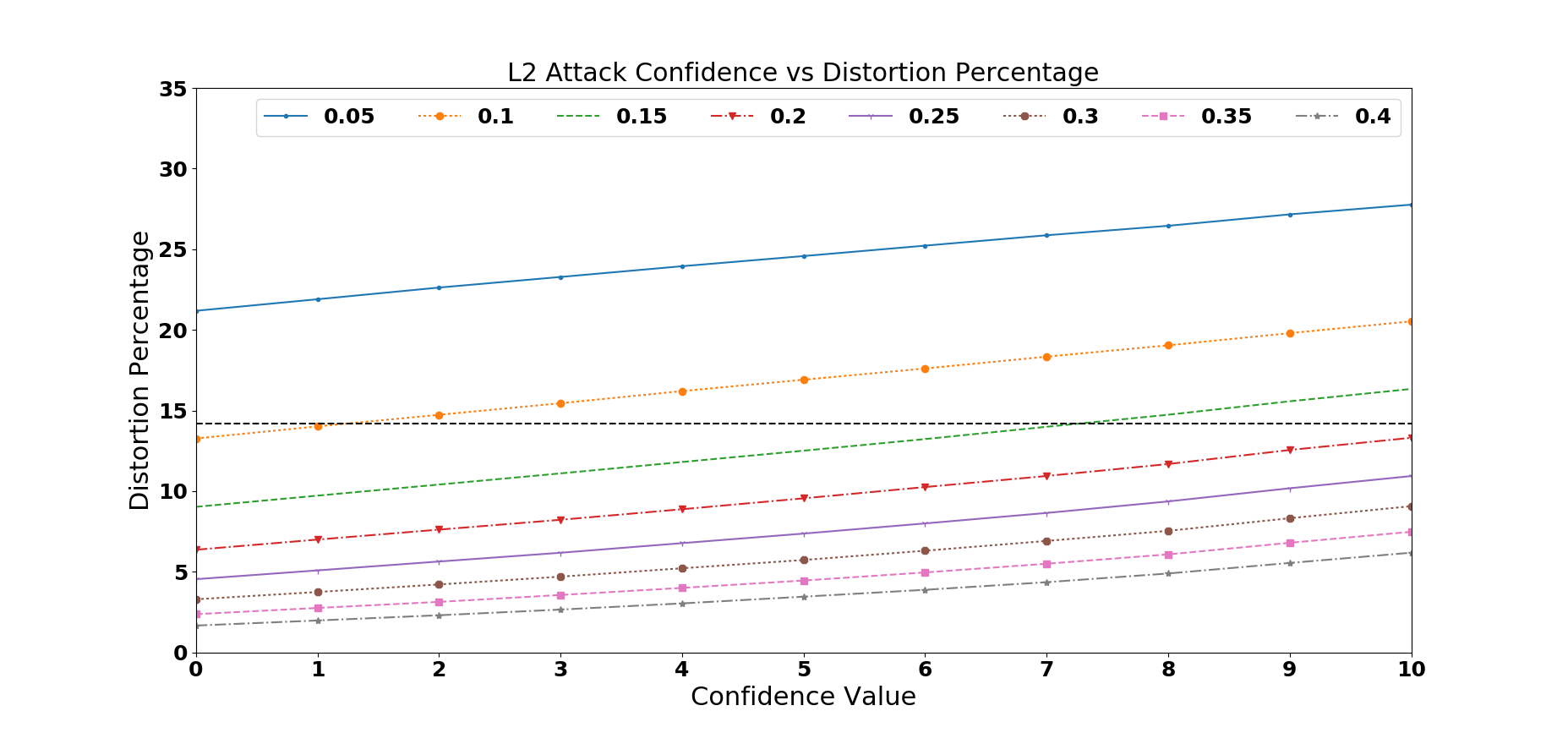}
\caption{Distortion comparison for attack confidence values of 0-10 on secure model. The dashed line at 14.2\% represents the threshold that humans will likely misclassify adversarial images.}
\label{fig:DistortionSecure}
\end{figure}

This result highlights the impact that dropout alone can have on the adversarial resistance of a model. But how does this visually impact a result to a human observer in this specific application? In section \ref{sec:compareRes} an analysis of the percentage of pixels visibly distorted in an image is presented on the comparison LeNet model. This same experiment is run on the securely trained model that includes dropout on the input layer, and the results are shown in figure \ref{fig:DistortionSecure}. A similar trend can be observed in both of these graphs, however the distortion level of the newly constructed model for greater values of minimum modification percentages increases slightly, while it decreases for values of 0.05 modification of 0.10.

Although the minimum modification percentage per pixel required for a human to notice a change is unknown, the increase in distortion for all minimum percentages above 0.10 is desirable, and signifies that this model may in fact be more secure. In fact, for each these minimum distortion values, the percent of pixels that are modified more than the minimum is less than 14.2\% in the original model, which is the value that \cite{papernot_thesis} identified as the percentage of modified pixels in an image that causes a massive increase in human misclassification of an image. Because these values are increasing to reach a distortion percentage of closer to 14.2\%, the model trained securely according the criteria previously discussed should be considered more secure than the LeNet model used for comparison in these experiments.

\begin{figure}
\centering
\includegraphics[width=3.25in,height=3.25in,keepaspectratio]{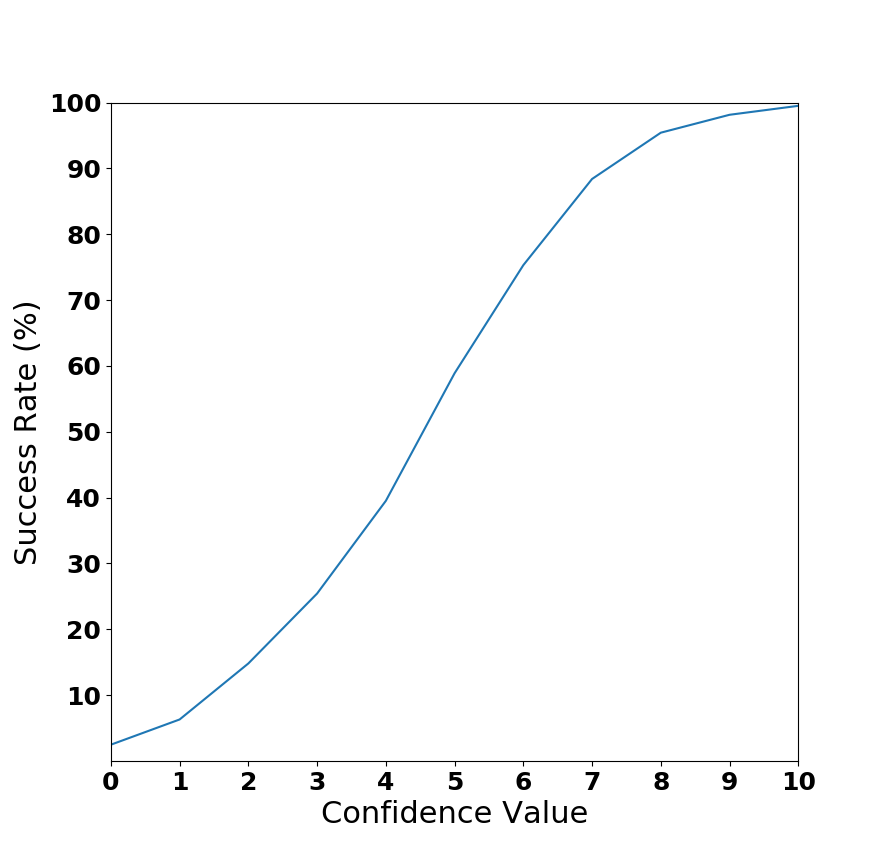}
\caption{The success rate of adversarial examples on the secure model which were originally crafted on the LeNet model drops far below 100\%, indicating an increase in adversarial resistance.}
\label{fig:accCompare}
\end{figure}

Another way to test whether this model is more secure than the original LeNet model is to measure the success rate of adversarial examples crafted on the original LeNet model on the more secure model. Samples that are created with low confidence are classified almost equally as the source class and the target class, so they represent the closest adversarial examples that exist on a model. Figure \ref{fig:accCompare} shows this comparison with adversarial samples crafted using attack confidence values between 0 and 10. The success rate is as low as 2\% on close samples that were created with a confidence level of 0 on the original model, and stays relatively low for mid confidence levels in the attack, which indicates that close adversarial examples that exist on the original model no longer exist on this model. 

\begin{figure}
\centering
\subfloat[Source Class: 7, Target Class: 0.]{
    \centering
    \includegraphics[width=\textwidth,height=\textheight,keepaspectratio]{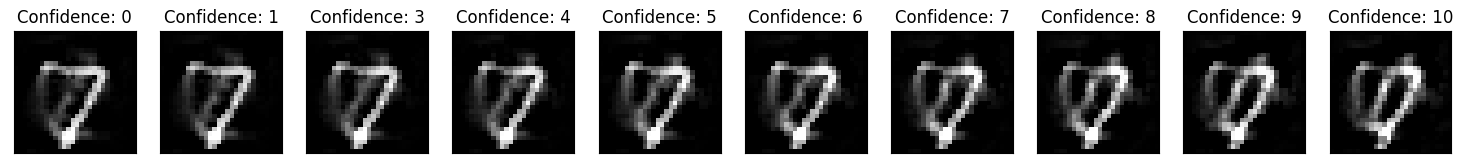}}
    \\
\subfloat{
    \centering
    \includegraphics[width=\textwidth,height=\textheight,keepaspectratio]{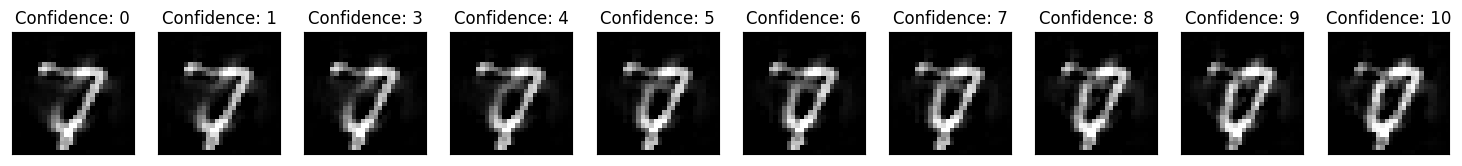}}
    \\
\subfloat[Source Class: 4, Target Class: 8.]{
    \centering
    \includegraphics[width=\textwidth,height=\textheight,keepaspectratio]{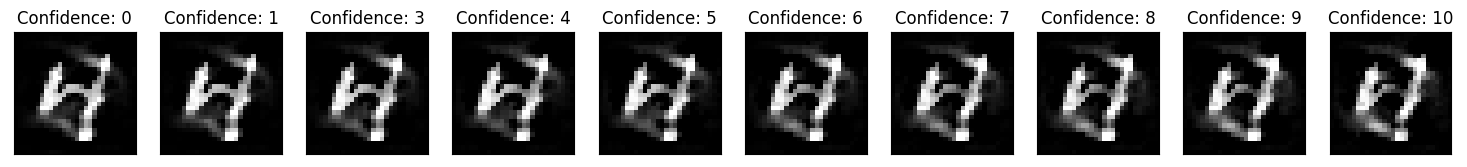}}\\
\subfloat{
    \centering
    \includegraphics[width=\textwidth,height=\textheight,keepaspectratio]{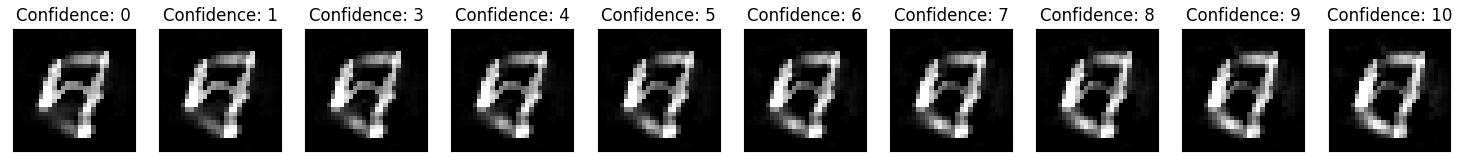}}
\caption{Visual comparison of distortion for secure and original model as confidence level increases.}
\label{fig:compareSecure}
\end{figure}

Finally, a visual comparison of the adversarial examples observed on both of these models is provided in \ref{fig:compareSecure}. As the attack confidence level increases, the images are more distorted to better resemble the target class. The top figures in each comparison are adversarial examples on the original LeNet model used for comparison in this work, and the bottom figures are samples constructed on the LeNet model with the addition of parameters that were identified to increase adversarial resistance. All of these images were successfully misclassified as the target class.

Notice that as the confidence level increases, it is more likely to visually identify the image as the target class instead of the source class. In a secure classifier, a human would recognize each of these images as only the target class, so visually the goal is to increase the salient features of the target class for all adversarial examples.

In both instances provided, the target class is more recognizable in the securely trained model than it is in the original model. As the confidence level increases, it seems far more likely to classify the sample as the target class than the source class in the secure model, suggesting that the additional secure parameters positively impact the resistance of the model to adversarial examples.

\section{Discussion}
\label{sec:discussion}

Although the choice of network hyperparameters alone is not sufficient to eliminate the impact of adversarial examples on a system, it does have an impact on a model's ability to prevent the existence of \textit{close} samples. The visual impact of parameters such as dropout on adversarial examples is significant, and could be enough to cause a human victim to correctly classify a sample that is intended to be malicious. It is also important to understand that not all parameters have a significant impact on the adversarial example generation on a model. For example, the number of hidden units and hidden layers had very little impact on adversarial resistance of a model.

So why do parameters such as pool size, kernel size, and dropout on the input layer impact adversarial resistance? One interesting observation that can be made about the results in this work is the correlation between overfitting and adversarial distance. The same parameters that are generally used to reduce the risk of overfitting, such as dropout and pool size, have a positive impact on the measured adversarial distance between classes. Intuitively, this makes sense because these parameters help the trained model generalize to testing time data that has not previously been observed, which is one way to view adversarial examples.

Another way to understand why these parameters have an impact on adversarial distance is the drastic change in features that are extracted depending on each of these parameters.  Specifically relating to dropout, if 50\% of the input is missing from an image, a large portion of that image is missing during the forward pass through the network. This process effectively expands the training data to include input that would not otherwise be observed in the environment, which is exactly how adversarial examples were previously described in this work.

Future research into adversarial example crafting techniques and defenses should consider the implications of models trained using different hyperparameters. When comparing attacks and defenses, the models that are attacked should be trained with the same values for higher impact hyperparameters such as kernel size, pool size, and dropout, and should use the same optimizer for training. Without using similar hyperparameters in training, the models being compared may have different levels of resistance to adversarial examples, so the results gathered from attacks or defenses against different models cannot be compared. Specifically for the MNIST dataset, our results provide hyperparameter values that can be used to train a more secure model, which should be used as a standard base model to compare against future defenses.

\subsection{Threat Modeling}

It is clear that adversarial examples likely \textit{exist} on a large majority of modern classifiers, but can these examples truly be exploited on a real system in a way that aids an attacker? While this is largely dependent on the information available to the attacker and the attacker's capability, this work shows that every data set has its own properties and vulnerabilities. Taking this into account, the following items are briefly discussed as important considerations for threat modeling a CNN.

\begin{itemize}

\item \textbf{Oracle}. Current adversarial example crafting techniques rely on iterative optimization processes to find adversarial examples. In each round of these iterative attacks, the adversary must be able to extract the output of the network on the current adversarial example. This feedback is called an \textit{oracle}, and is essential for an attacker to reliably craft adversarial examples on a network. Systems that do not provide a strong oracle for an attacker are more difficult to optimize against. Further research on this subject may provide insight into the impact that oracle strength has on the strength of adversarial examples.

\item \textbf{Intra-Class Distance}. In this work, it is shown that some source-target class pairs are more vulnerable to attack than others. Understanding these patterns for the data type in an application is essential for understanding the threats to that application; some classes may be more easy to target than others.

\item \textbf{Critical Classes}. In some applications, such as authorization applications, specific classes may be more critical than others. For instance, high privileged administrative users may be more frequently targeted than lower privileged or anonymous users. In this case, the adversarial distance between a critical class and all other classes should be maximized more than the distance between less critical classes. If a low privileged user can be classified as a high privileged user with only a small change in behavior, the model is not very effective.

\end{itemize}

\subsection{Defense Against Adversarial Examples}

The issue with adversarial training as a defense is the computational and storage requirements for creating and retaining all possible adversarial examples on a model. If this process can be automated in the model training process without the explicit need to craft an adversarial example, the same success could likely be had without the storage and computation problems. 

One approach that would be interesting would be to harness dropout in more advanced way that effectively produce these samples. More advanced, application specific designs for these dropout layers may provide a means expand the types of features that are learned in each class without adding additional training data. For example, if samples that represent a seven are passed through a dropout-like layer that only drops out pixels which are spatially important for that sample to represent a seven, there may be a greater impact on the resistance of the resulting model to attack.

Another approach would be to craft a dropout-like layer which instead of just dropping out some features also intelligently modifies some feature to automatically add noise to images. Many times, the additions made to an image with an l2 attack that to cause it to be misclassified simply resemble noise, and any additional noise that is added to an image may increase a model's ability to represent that additional noise correctly. 

Finally, an application level approach could also be taken to defend this type of attack. If all data is cleaned by the application to fit a specific \textit{class template} before being classified, some of these attacks may be easily prevented. For example, if a faint circle is added to a 7 in MNIST to cause it to be misclassified as a 0, an application could change the value of any pixel that is less than 200 to become 0. This would effectively delete the attack, which is the faint circle, and the resulting classification would be a 7.

\section{Conclusion}
\label{sec:conclusion}

This paper has explored the ability of various hyperparameters to impact the adversarial resistance of a model. This is an important subject for researchers that are exploring adversarial examples to understand while they are crafting new attacks and defense techniques. If a model used in one paper is trained with dropout on the input layer while a competing work does not use dropout, the significance of the attacks should not be directly compared because of the impact that dropout can have on adversarial resistance of the model. 

The contributions of this work include the following; first, a new analysis of MNIST is presented based on modern evasion attacks. This analysis is used to craft samples that can be used to estimate the vulnerability of a model to adversarial examples while under computational constraints. Following this, a variety of hyperparameters are experimented on to determine which parameters most contribute to adversarial resistance. This is done with the samples that were found using the MNIST evaluation, and resulted in the discovery that dropout, pool size, and kernel size were some of the main contributors to adversarial resistance of a model. A "secure" model is crafted with these parameters, and a comparison of the visual impact that these parameters have on adversarial examples is done with the secure model and a basic LeNet model to determine if they truly cause a humanly visible impact, which they do.

The main findings are that dropout on the input layer, the ReLU activation unit, larger kernel sizes, smaller pool sizes, and stochastic gradient descent are the main contributors to adversarial resistance in a model. While they do indeed help a model generalize to adversarial examples when used together, they currently should not be used in and of themselves as viable defense mechanisms.

\subsection{Future Work}

While the results in this work shed light on the impact of commonly used hyperparameters on the security of convolutional neural networks, there are many design decisions that are not considered in scope of this work. Some of these decisions may yield interesting results. For example, how is the security of CNNs impacted when multiple hyperparameters are changed in conjunction with each other? Do different deep learning architectures and datasets respond differently to evasion attacks? How do these training decisions and hyperparameters impact the success of adversarial training? These questions remain unanswered in this work, and are a provocative direction for further research into this area.

Some of the parameters tested in this work generated curious results, such as the lack of impact on adversarial distance of the number of dense nodes and number of hidden layers in a network. If there is a correlation between overfitting and adversarial distance between classes, these hyperparameters should have had a more substantial impact on the attack. Another interesting observation in this work is the unexpected impact of different optimizers on the measured adversarial distance. Further research into these results may provide deeper insights into adversarial examples and the security of CNNs.

Finally, the success of dropout on the input layer of a network is promising as a potential defense mechanism. While dropout is shown to have a small amount of success at increasing the adversarial distance between classes, more advanced dropout layers that can intelligently select input features to drop might have a more substantial impact on the ability of a model to generalize to adversarial examples.


\bibliography{main}


\appendix
\newpage
\appendix
\section*{Appendix}
\label{Appendix}

\begin{table}[H]
\centering
\caption{Specific parameters tested.}
\label{fig:allspecificParams}
\begin{tabular}{|c|c|}
\hline
Parameter  & Specific \\
\hline
Activation Units & ReLU\\
\cline{2-2}
&Sigmoid\\
\cline{2-2}
&Linear\\
\cline{2-2}
&Tanh\\
\hline
Dropout on Input Layer & 0.1\\
\cline{2-2}
&0.2\\
\cline{2-2}
&0.3\\
\cline{2-2}
&0.4\\
\cline{2-2}
&0.5\\
\hline
Dropout on Dense Layer & 0.1\\
\cline{2-2}
&0.2\\
\cline{2-2}
&0.3\\
\cline{2-2}
&0.4\\
\cline{2-2}
&0.5\\
\hline
\makecell{Convolution Kernel Size\\(Dimensions)} & (2,2)\\
\cline{2-2}
&(3,3)\\
\cline{2-2}
&(4,4)\\
\cline{2-2}
&(5,5)\\
\cline{2-2}
&(6,6)\\
\hline
\end{tabular}
\end{table}

\begin{table}[]
\centering
\caption{Specific parameters tested(contiued).}
\label{fig:allspecificParams2}
\begin{tabular}{|c|c|}
\hline

\makecell{Max Pooling Size\\(Dimensions)}  & (2,2)\\
\cline{2-2}
&(3,3)\\
\cline{2-2}
&(4,4)\\
\cline{2-2}
&(5,5)\\
\cline{2-2}
&(6,6)\\
\hline
Weight Initializers & Orthogonal\\
\cline{2-2}
&Truncated Normal\\
\cline{2-2}
&Random Uniform\\
\cline{2-2}
&Glorot Uniform\\
\cline{2-2}
\hline
Bias Initializers & Random Uniform\\
\cline{2-2}
&Truncated Normal\\
\cline{2-2}
&zeros\\
\hline
Number of Hidden Units & 100\\
\cline{2-2}
&200\\
\cline{2-2}
&300\\
\cline{2-2}
&400\\
\cline{2-2}
&500\\
\hline
Number of Dense Layers & 1\\
\cline{2-2}
&2\\
\cline{2-2}
&3\\
\cline{2-2}
&4\\
\cline{2-2}
&5\\
\hline
Number of Filters per CNN & 32/64\\
\cline{2-2}
&64/128\\
\cline{2-2}
&128/256\\
\cline{2-2}
&256/512\\
\cline{2-2}
&512/1024\\
\hline
\end{tabular}
\end{table}

\begin{table}[]
\centering
\caption{Specific parameters tested(continued).}
\label{fig:allspecificParams3}
\begin{tabular}{|c|c|}
\hline

Kernel Regularizers & L1\\
\cline{2-2}
&L2\\
\cline{2-2}
&L1 and L2\\
\hline
Avtivity  Regularizers & L1\\
\cline{2-2}
&L2\\
\cline{2-2}
&L1 and L2\\
\hline
Batch Size & 32\\
\cline{2-2}
&64\\
\cline{2-2}
&128\\
\cline{2-2}
&256\\
\cline{2-2}
&512\\
\hline
Loss & Log Loss\\
\cline{2-2}
&Hinge Loss\\
\cline{2-2}
&MSE\\
\cline{2-2}
&Cross Entropy\\
\cline{2-2}
&Sigimoid Cross Entropy\\
\hline
Optimizers & Stochastic Gradient Descent\\
\cline{2-2}
&Adadelta\\
\cline{2-2}
&Adagrad\\
\cline{2-2}
&Adam\\
\cline{2-2}
&Adamax\\
\cline{2-2}
&Nadam\\
\cline{2-2}
&RMSProp\\
\hline
Momentum & 0.2\\
\cline{2-2}
&0.4\\
\cline{2-2}
&0.6\\
\cline{2-2}
&0.8\\
\cline{2-2}
&1.0\\
\hline

\end{tabular}
\end{table}
\raggedbottom\sloppy

\end{document}